%% file: root.tex
\documentclass[letterpaper, 10 pt, conference]{IEEEtran}  

\IEEEoverridecommandlockouts                              

\usepackage{amsmath,amssymb,amsfonts}
\usepackage{algorithm}
\makeatletter
  \renewcommand{\ALG@name}{Workflow}
\makeatother
\usepackage{algpseudocode}
\usepackage{booktabs}
\usepackage{graphicx}
\usepackage{subfig}
\usepackage{mathtools}
\usepackage{fancyhdr}

\title{\LARGE \bf
 Motion Compensation for Real Time Ultrasound Scanning in Robotically Assisted Prostate Biopsy Procedures*
}

\author{Matija Markulin$^{1}$, Luka Matijević$^{1}$, Luka Šiktar$^{1}$, Janko Jurdana$^{1}$, \\ Branimir Ćaran$^{1}$, Marko Švaco$^{1}$, Filip Šuligoj$^{1}$ and Bojan Šekoranja$^{1}$
\thanks{*This project has been funded by the European Union – NextGenerationEU through the Recovery and Resilience Facility.}
\thanks{$^{1}$Authors are with Faculty of Mechanical Engineering and Naval Architecture,
        University of Zagreb, Ivana Lučića 5, 10000 Zagreb, Croatia
        {\tt\small matija.markulin@fsb.unizg.hr}}%

}

\begin{document}

\maketitle
\thispagestyle{fancy}
\fancyhf{}
\renewcommand{\headrulewidth}{0pt}
\fancyhead[C]{Preprint submitted for ICRA 2026}
\pagestyle{fancy}
\fancyhf{}
\renewcommand{\headrulewidth}{0pt}
\fancyhead[C]{Preprint submitted for ICRA 2026}

\begin{abstract}
Prostate cancer is one of the most common types of cancer in men. Its diagnosis by biopsy requires a high level of expertise and precision from the surgeon, so the results are highly operator-dependent. The aim of this work is to develop a robotic system for assisted ultrasound (US) examination of the prostate, a prebiopsy step that could reduce the dexterity requirements and enable faster, more accurate and more available prostate biopsy. We developed and validated a laboratory setup with a collaborative robotic arm that can autonomously scan a prostate phantom and attached the phantom to a medical robotic arm that mimics the patient's movements. The scanning robot keeps the relative position of the US probe and the prostate constant, ensuring a consistent and robust approach to reconstructing the prostate. To reconstruct the prostate, each slice is segmented to generate a series of prostate contours converted into a 3D point cloud used for biopsy planning. The average scan time of the prostate was 30 s, and the average 3D reconstruction of the prostate took 3 s. We performed four motion scenarios: the phantom was scanned in a stationary state (S), with horizontal motion (H), with vertical motion (V), and with a combination of the two (C). System validation is performed by registering the prostate point cloud reconstructions acquired during different motions (H, V, C) with those obtained in the stationary state. ICP registration with a threshold of 0.8 mm yields mean 83.2\% fitness and 0.35 mm RMSE for S-H registration, 84.1\% fitness and 0.37 mm RMSE for S-V registration and 79.4\% fitness and 0.37 mm RMSE for S-C registration. Due to the elastic and soft material properties of the prostate phantom, the maximum robot tracking error was 3 mm, which can be sufficient for prostate biopsy according to medical literature. The maximum delay in motion compensation was 0.5 s. Our approach can compensate for movements in a plane perpendicular to the probe axis.

\end{abstract}

\begin{IEEEkeywords}
robotics, motion compensation, ultrasound, machine learning, prostate biopsy
\end{IEEEkeywords}


\input{Text/1_Intro}
\input{Text/2_Method}
\input{Text/3_Result}
\input{Text/4_Conclusion}






\section*{ACKNOWLEDGMENT}

We would like to thank dr. Kuliš and prof. Hudolin and the whole team from the Clinical Hospital Center Zagreb for providing us with data and expertise for the training of the neural network.

We would like to acknowledge that Figure \ref{fig:scanning} was created with the help of ChatGPT.


\bibliographystyle{IEEEtran}
\bibliography{Litereature}

\end{document}

%% file: Text/1_Intro.tex
\section{Introduction}

Prostate cancer is one of the most common cancers in men \cite{noauthor_globocan_nodate}, and its diagnosis is based on the histopathological evaluation of prostate biopsy samples. The procedure requires a high level of skill on the part of the clinician performing it, and the results are highly operator-dependent \cite{barrett_quality_2023}. Therefore, it is crucial to make the process less complex and reduce the skill requirements to ensure better patient care. A robotic system can address this challenge, as recent studies have shown that robotic assistance provides more precise biopsy needle positioning compared to the manual approach \cite{tilak_3t_2015}, \cite{lim_robotic_2019}.

Since targeted prostate biopsy has higher success rates than systematic biopsy \cite{porpiglia_multiparametric-magnetic_2016}, a novel approach that enables the surgeons to more easily plan the prostate biopsy locations is needed. There are already some implementations of robotically assisted prostate biopsy that combine transrectal ultrasound (TRUS) and magnetic resonance imaging (MRI). These systems usually determine the needle insertion point and depth while the surgeon inserts and fires the needle gun \cite{wetterauer_diagnostic_2021,maris_toward_2021}. One of the leading systems in this field is the iSR’obot Mona Lisa™ from Biobot Surgical \cite{ho_robotic_2011}, which is already being used in clinical practice. Furthermore, various types of robot kinematics and approaches for transperineal prostate biopsies have been developed and tested \cite{zang_7_dof,duan_continous_robot,wang_mri_trus}. However, there are still no reports of robotic systems with active motion compensation that meet the high accuracy requirements for ultrasound (US) prostate scanning.

One of the biggest challenges in prostate scanning and biopsy is a robust implementation of control loops and safety measures while the robot is working in close proximity to the human body. In \cite{ipsen_towards_2021}, the authors demonstrate autonomous ultrasound image acquisition over long periods of time with motion compensation for pseudoperiodic movements. Another similar approach is described in \cite{suligoj_robustautonomous_2021}, where a robot performs an autonomous US scan of the jugular vein. The aforementioned Mona Lisa™ system does incorporate some form of motion compensation, but relies primarily on software-based adjustments of ultrasound slices that are affected by motion-induced forces. In contrast, prostate biopsy procedure could benefit significantly from an active motion compensation system that is able to follow the movement of the prostate directly and in real time. It has the potential to reduce the cognitive burden on clinicians by eliminating the need to mentally reconstruct the position of the prostate \cite{Deng_Liu_Wang_Ruan_Li_Wu_Qiu_Wu_Tian_Yu_et_al._2024, Liu_Xiang_Xu_Zhang_Xu_2022}.

In order to automate the process, the system must be able to perform autonomous segmentation of the prostate. The MicroSegNet \cite{jiang_microsegnet_2024} model is the current state-of-the-art approach in Micro-Ultrasound segmentation. It is based on TransUNet \cite{chen_transunet_2021}, a variation of U-Net \cite{ronneberger_u-net_2015}, to create a reliable segmentation mask. This model was used in \cite{robotics14080100} for phantom segmentation in robotically assisted prostate biopsy.

Our robotic system reduces the expertise and dexterity required by clinicians and contributes to more consistent procedural outcomes. To address this challenge, we developed a two-robot setup that mimics the real-world scenario. One robot holds the prostate phantom and mimics the human movements that may occur during the procedure, while the other holds an ultrasound probe with a biopsy targeting device and performs prostate scanning and biopsy treatment. To compensate for the movements of the patient that the robotic arm makes when moving the phantom, we have developed a system that provides compensation in two different ways. Perpendicular motions relative to the ultrasound probe’s axis are corrected by using force control, while it adjusts the initial position along the length of the probe based on the US image.

With the proposed solution, we aim to enable scanning with constant force on the tissue. The constant force is crucial for a smooth scanning process without excessive tissue deformation. Throughout the scannning process, segmentation is performed on the acquired ultrasound images. The stored ultrasound images and their segmentation results are used for prostate reconstruction. A correct reconstruction of the prostate ultrasound scan, despite possible patient movement, can only be achieved by utilizing precise movement compensation. Our system not only keeps the relative position of the robot and phantom constant, but also constantly monitors the applied forces and interrupts the scanning process if the force is outside the required range. This ensures a consistent and robust approach to reconstructing the prostate, which is explained in detail in Workflow \ref{workflow_biopsy}.

%% file: Text/2_Method.tex
\section{Methodology} 

\begin{figure}[h]
    \centering
    \includegraphics[width=\linewidth]{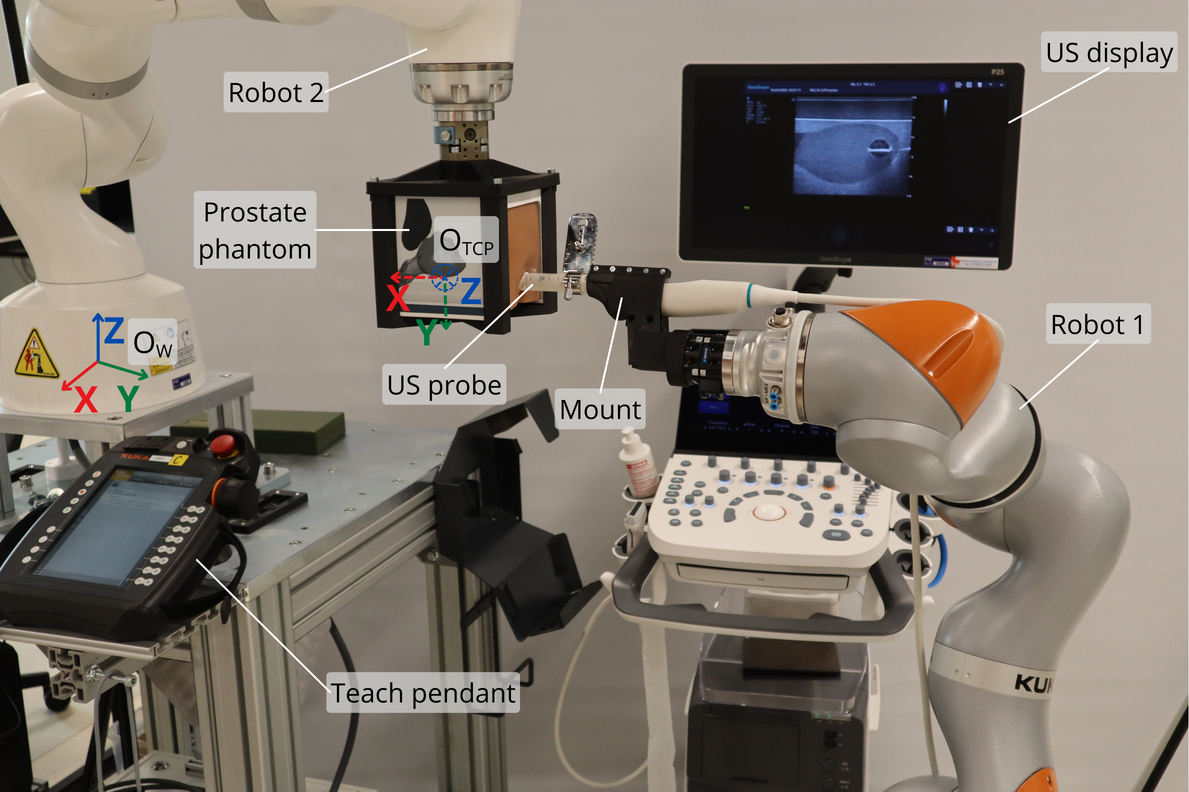}
    \caption{Robot setup: A robot-mounted ultrasound probe is inserted into a robot-mounted prostate phantom. By rotating the probe around its x-axis, the robot can scan the entire prostate.}
    \label{fig:setup}
\end{figure}

The robot setup consists of two KUKA LBR robots, the industrial variant, IIWA (robot 1 in Figure \ref{fig:setup}) and the medical variant LBR Med (robot 2). The KUKA IIWA was used as a surgical assistant with an ultrasound probe attached to it, as it was easier to assemble when emulating a future clinical setup proposed by experienced urologists. The Med robot has the connectors for the power and data cables on the inside of the mounting flange, which would require a redesign of the robot table. The robot is placed on a low table to take up as little space as possible in the clinician's working area and to make the biopsy workflow as similar as possible to the traditional method.

\begin{algorithm}[h]
    \caption{Robotic Prostate Phantom Biopsy Simulation}\label{workflow_biopsy}
    \begin{algorithmic}[1]
    \State Insert probe into phantom (hand-guidance mode)
    \State Robot~2 starts random anthropomorphic movements within biopsy range
    \State Robot~1 monitors contact force between probe and phantom and applies compensatory movements to maintain stable probe–phantom contact

    \State \textbf{Scanning procedure:}
        \Statex \hspace{\algorithmicindent} Robot~1 starts ultrasound imaging sweep, applies MicroSegNet segmentation model with additional classification head
        \Statex \hspace{\algorithmicindent} If Robot~2 induces motion $\rightarrow$ Robot~1 detects excessive force
        \Statex \hspace{\algorithmicindent} Stop scanning immediately at the current slice, compensate motion to remain aligned with stopped slice
        \Statex \hspace{\algorithmicindent} Resume scanning once Robot~2 stabilizes

    \State \textbf{Reconstruction:}
        \Statex \hspace{\algorithmicindent} Extract segmented contours into 3D point cloud $\rightarrow$ reconstruct full prostate 
    \State \textbf{Biopsy procedure:}
        \Statex \hspace{\algorithmicindent} Rotates to the specific target slice for needle insertion, also compensates motion 
    \end{algorithmic}
\end{algorithm}

The robot is controlled over the Robot Operating System 2 (ROS2) platform using open source support packages \textit{lbr-stack} \cite{Huber2024}. The robot is programmed using Python for easier integration with computer vision and Machine Learning libraries. Specifically, our solution uses Numpy and Scipy for working with matrices, OpenCV for image preparation, and CUDA and PyTorch for running the segmentation.

A SonoScape P25 ultrasound (US) system with the BCL10-5 biplanar transrectal ultrasound probe (TRUS) is mounted on the flange of the robot 1. The TRUS probe's linear US element produces an image 60 mm wide and between 3 and 90 mm deep, defined in the US image settings. The TRUS probe emits a 7.5 MHz signal. Our images were taken from a received frequency of 5.5-10 Hz, with a dynamic range of 120, a focus at 30 mm depth and an overall depth of 50 mm. The US image frames are streamed to a PC for real-time analysis and storage.

To simulate the movement of the patient in a controlled and measurable fashion, the prostate phantom was mounted on another robot. For this purpose, we used the Med variant of the KUKA LBR robot (robot 2 in Figure \ref{fig:setup}). A robot movement algorithm, simulating the actual human movements during the procedure, in the range of 15 mm, is created using ROS2.

\begin{figure}[h]
    \centering
    \includegraphics[width=\linewidth]{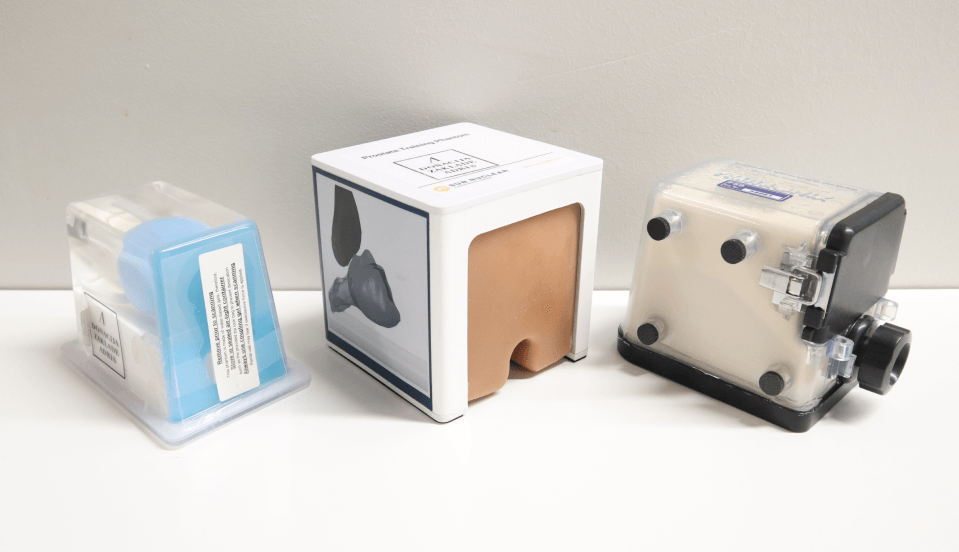}
    \caption{Medical prostate phantoms. From left to right: CIRS 053L, CIRS 070L, Yezitronix S-MM-3.1.}
    \label{fig:phantoms}
\end{figure}

Throughout our research, multiple prostate phantoms with different credibility, compared to the real prostate, were tested in a robot setup. Models 053L and 070L, made by CIRS and S-MM-3.1 by Yezitronix (Figure \ref{fig:phantoms}), were tested for the motion compensation algorithm. The 053L showed the least credibility, thus unable to be used for our use case. CIRS 070L proved the best as the insertion area is the most similar to the real one, and the prostate phantom represents the real prostate. The S-MM-3.1 phantom has a big insertion area and the cutout on the plastic casing is not aligned properly. Also, this phantom is not firmly fixed inside the casing, thus it cannot be tested using our setup.

The motion of the phantom was performed with the following speed profile:
\begin{equation*}
v = 
     \begin{cases}
       \cos{\frac{t}{2}} \quad & \frac{t}{2\pi} - \lfloor{\frac{t}{2\pi}}\rfloor > 0.75 \\
       0 & \text{otherwise} \\

     \end{cases}
\end{equation*}
This speed profile simulates short, involuntary patient movements triggered by discomfort during the procedure. In the case of combined vertical and horizontal motions, the two profiles are identical, with the horizontal motion phase shifted by one-eight of the phase relative to the vertical.

\section{Scanning, Segmentation and reconstruction}\label{sec:segment}
\begin{figure}[h]
    \centering
    \includegraphics[width=\linewidth]{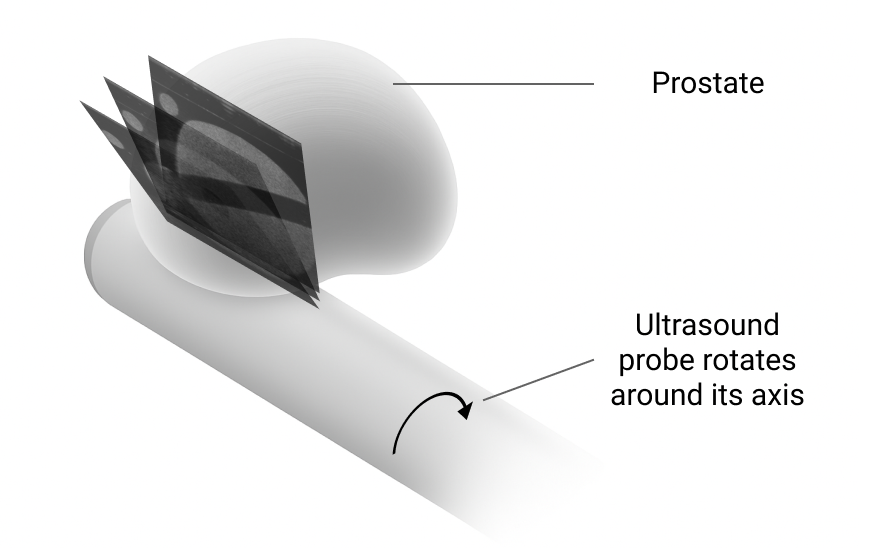}
    \caption{Ultrasound scanning procedure. The robot rotates the probe around its axis to acquire images (slices) of the whole prostate.}
    \label{fig:scanning}
\end{figure}

The ultrasound scanning (sweep) is performed by rotating the US probe around its x-axis from one end of the prostate to the other. The robot rotates the probe while compensating for the phantom's movement to scan the whole prostate. Figure \ref{fig:scanning} illustrates the scanning process.

\begin{figure}[ht]
    \centering
    \subfloat[\centering]{\includegraphics[width=0.49\linewidth]{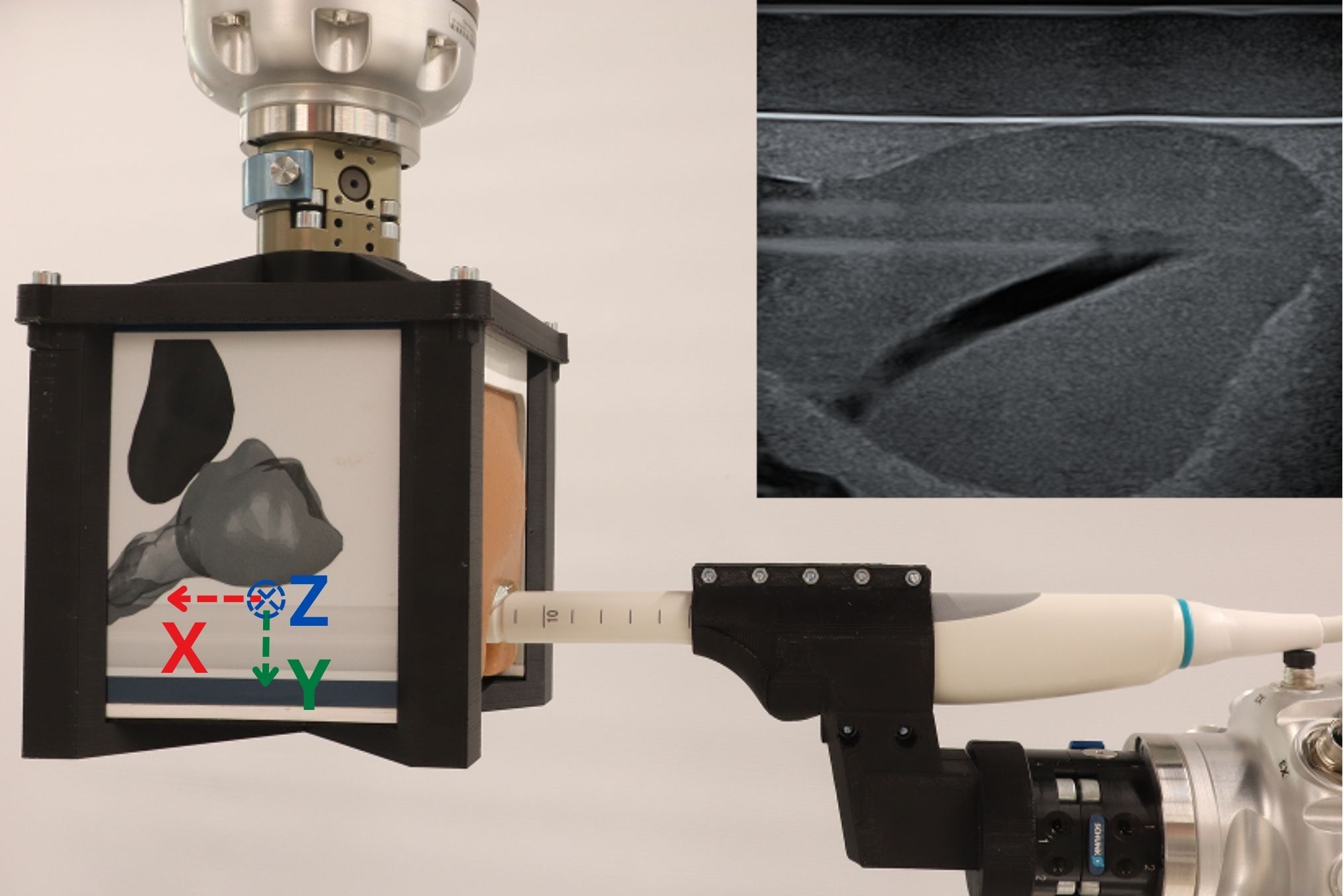}}
    \hspace{1 px}
    \subfloat[\centering]{\includegraphics[width=0.49\linewidth]{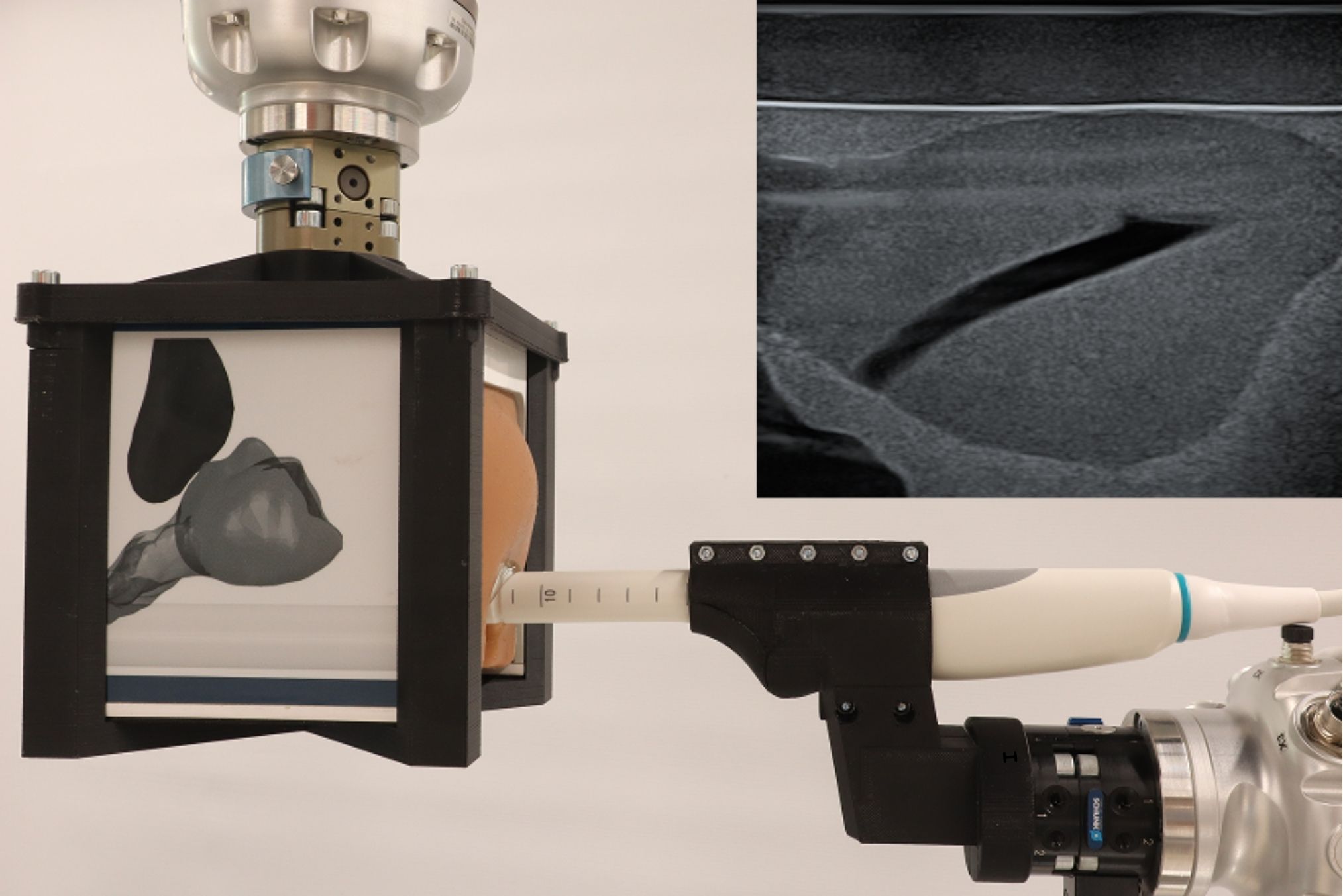}}
    \caption{Prostate phantom scan during \textbf{(a)} stationary state (without deformations) and \textbf{(b)} downward motion of the phantom (visible deformations)}
    \label{fig:Phantom_deformation}
\end{figure}

First, the US probe is placed in the middle of the prostate, such that the urethra is in view, using hand-guidance mode. From there, the robot rotates the probe to one side until the edge is detected by the segmentation model and the robot changes the direction. The probe is then rotated to the other end while storing the US pictures and the corresponding segmentations, as well as position information, for the reconstruction. While the robot moves the soft tissue of the phantom deforms as shown in Figure \ref{fig:Phantom_deformation}, and to preserve the quality of the reconstruction, the robot pauses the sweep until the force returns below the threshold. The procedure is shown in Workflow \ref{workflow_segmentation}.

\begin{algorithm}[h]
    \caption{Segmentation and Reconstruction}\label{workflow_segmentation}
    \begin{algorithmic}[1]
    \Require threshold
    \State \textbf{Sweep}
    \State Place probe in initial position with hand-guidance mode
    \State Start motion compensation
    \While{Prostate in sight}
    \State \hspace{\algorithmicindent} Rotate probe with speed 0.1 rad/s
    \EndWhile
    \State Change prostate direction
    \While{Prostate in sight}
    \If{$|F|>$ treshold}
    \State Wait for $|F|<$ treshold
    \EndIf
    \State Perform segmentation
    \State Save US picture and segmentation
    \State Rotate probe with speed -0.1 rad/s
    \EndWhile
    \State Stop and wait for biopsy locations
    \State \textbf{Reconstruction:}
    \For{Slices}
    \State Transform: image frame$ \rightarrow$ reconstruction frame
    \State Stack slice in reconstruction frame
    \EndFor
    \State Display reconstruction
    
    \end{algorithmic}
\end{algorithm}

\begin{figure}[h]
    \centering
    \includegraphics[width=\linewidth]{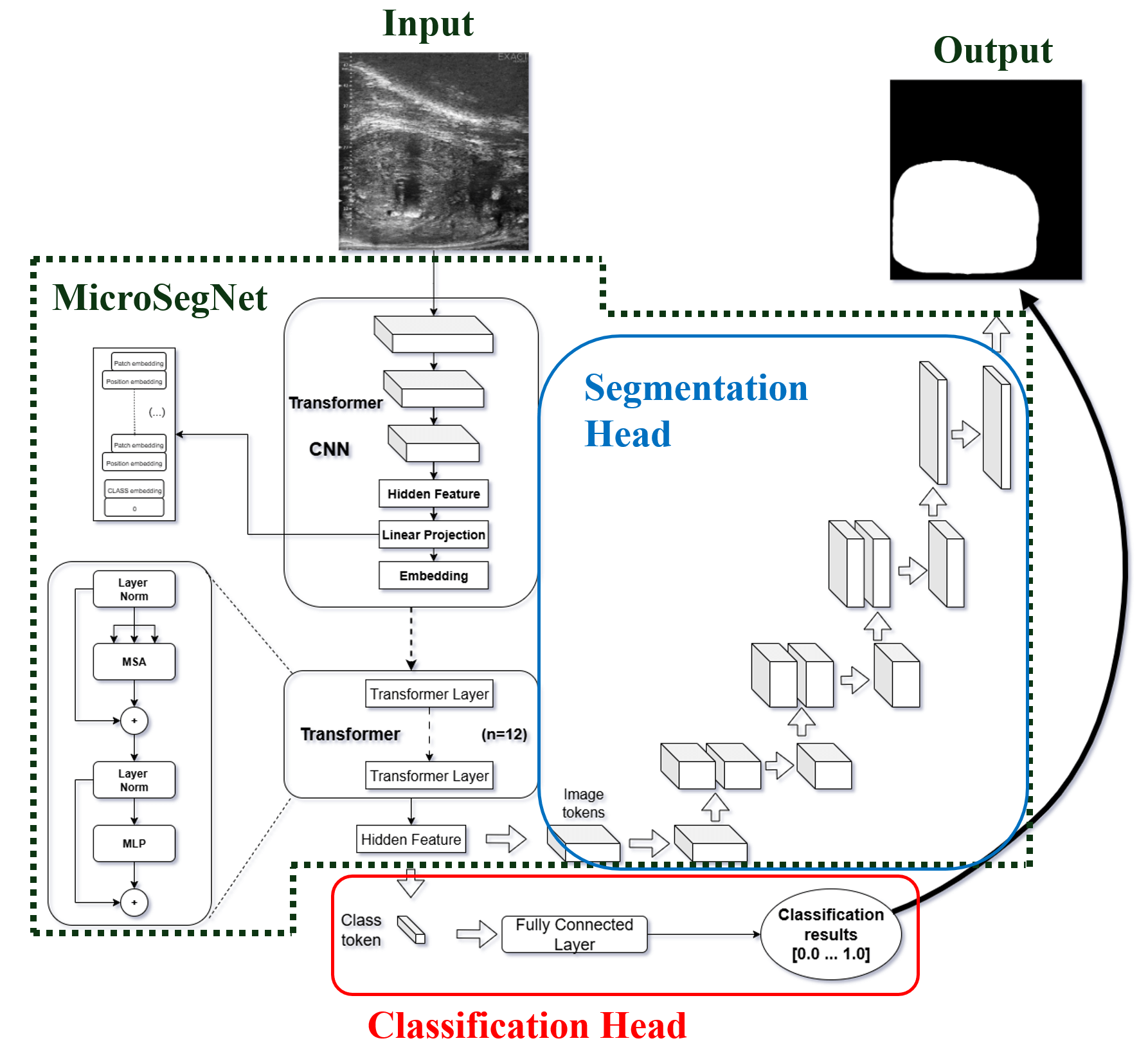}
    \caption{MicroSegNet with added classification head}
    \label{fig:microsegnet}
\end{figure}

Prostate segmentation processes the real-time acquired ultrasound images and outputs the segmentation masks, which are later used to extract the segmentation contours. The segmentation procedure uses MicroSegNet, a hybrid CNN-ViT segmentation model with an added classification head to detect the edges of the prostate, shown in Fig \ref{fig:microsegnet}. The CNN-ViT combination enhances the advances of both modalities. CNN model is used in this approach to extract local features from the input image, and ViT model is used to extract global, context information about the image. The model is fine-tuned on our own prostate phantom dataset, recorded especially for this research. Upon the successful fine-tuning on our dataset, the model was found to perform with an accuracy comparable to state-of-the-art ultrasound segmentation models, but has outliers at the edges. To overcome this problem, we introduced a novel classification head next to the segmentation head at the end of the ViT model, in order to classify the existence of the prostate on the ultrasound image. If the prostate is detected on the image, then the segmentation is performed, otherwise, the image is classified as a background with an empty segmentation mask.

The resulting prostate contours are used to create a point cloud representation of the prostate. The robotic arm provides the probe's angular position synchronized with each acquired ultrasound frame. The 3D reconstruction is performed by transforming each segmented slice from its 2D coordinate frame to the 3D visualization frame using the transformation matrix \textbf{T}. The visualization frame is aligned with the initial slice frame, which is located in the center of the prostate, in accordance with standard probe placement in clinical settings. The symbol $\phi$ denotes the angular offset of the probe from the central slice, and $r$ represents the radius of the probe of 9 mm, which is defined as the distance between the center of rotation (center of the US probe) and the beginning of the image acquisition area (surface of the US probe where the imaging element is located).

\begin{equation}
\label{eq:matrix_transformation}
\mathbf{T}
=
\begin{bmatrix}
1 & 0 & 0 & 0\\
0 & \cos(\phi) & -\sin(\phi) & \mathrm{r} \cdot \cos(\phi)\\
0 & \sin(\phi) & \cos(\phi) & \mathrm{r} \cdot \sin(\phi)\\
0 & 0 & 0 & 1
\end{bmatrix}
\end{equation}

\section{Methodology of motion tracking}

\subsection{Force control}
KUKA LBR robots measure the torque in each joint and calculate the part of these torques that is caused by external forces and torques from the robot's dynamic model. These external torques are then used via lbr-stack to estimate the forces and torques acting on the TCP, expressed in the TCP coordinate frame.

We use the estimated external forces and torques to program the high-level PID controller for keeping constant forces $F_y$ and $F_z$ (expressed in $O_{TCP}$ from Figure \ref{fig:setup} ) on the US probe. The referent value for $F_y$ is 7 N and for $F_z$ is 0 N. We have empirically tuned the controller to achieve fast tracking of soft phantom tissue. To prevent slight oscillations that occur without movement of the phantom, due to soft phantom material, a deadzone of 0.1 N is added to the controller. The external forces acting on the TCP are represented in the TCP coordinate frame, while the robot is controlled using desired speeds along the TCP axes through MoveIt’s servo commands \cite{Coleman_Sucan_Chitta_Correll_2014}.


\subsection{Visual positioning}
Force control is not used to adjust the position along the x-axis of the probe, since the force in the x-axis direction is caused only by friction and depends strongly on the scan parameters, such as the amount of US gel used. Instead of force control, we implemented a visual positioning approach for the control along the x-axis of the probe. The probe is centered on the prostate using the segmentation described in section \ref{sec:segment}. By calculating the center of mass for the segmentation mask at the specific slice, it is possible to calculate the distance between the US image center and the center of segmented prostate slice. The robot then moves the US probe to place the prostate in the center of the image.

While this approach reliably follows the motion, there is the limitation of the proposed positioning. The prostate is larger on one end compared to the another, which shifts the center of the segmentation back without any movement while scanning the prostate. Due to center shift, the robot then tries to compensate it, thus creating distorted reconstructions. Until this problem is solved, our approach uses visual positioning to set a consistent initial position for the sweeps, assuming no movement in this direction during the sweep.

%% file: Text/3_Result.tex
\section{Results}
To obtain statistically meaningful results, 30 sweeps for each of the four prostate phantom movement scenarios were recorded. The phantom was scanned in a stationary state (in Tables \ref{Tab:haus} and \ref{Tab:icp_table}, marked 'S'), with horizontal motion ('H'), with vertical motion ('V'), and with a combination of both horizontal and vertical movements ('C'). This setup enabled a systematic performance evaluation of our system under different motion conditions and the assessment of its robustness and accuracy in each scenario.

\subsection{Motion compensation}
\begin{figure}
    \centering
    \subfloat[\centering]{\includegraphics[width=\linewidth]{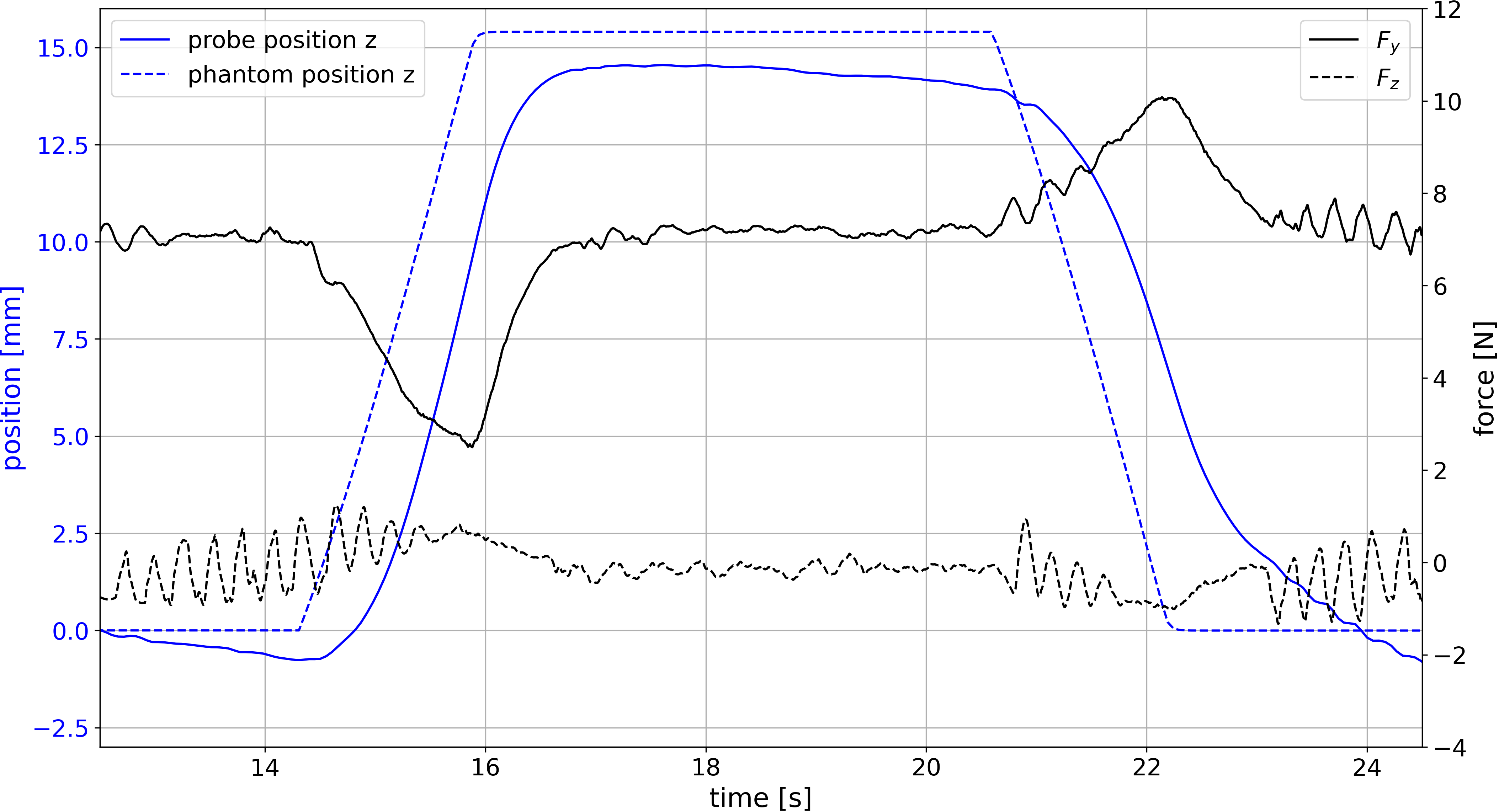}\label{fig:vert_mot}}\\
    \subfloat[\centering]{\includegraphics[width=\linewidth]{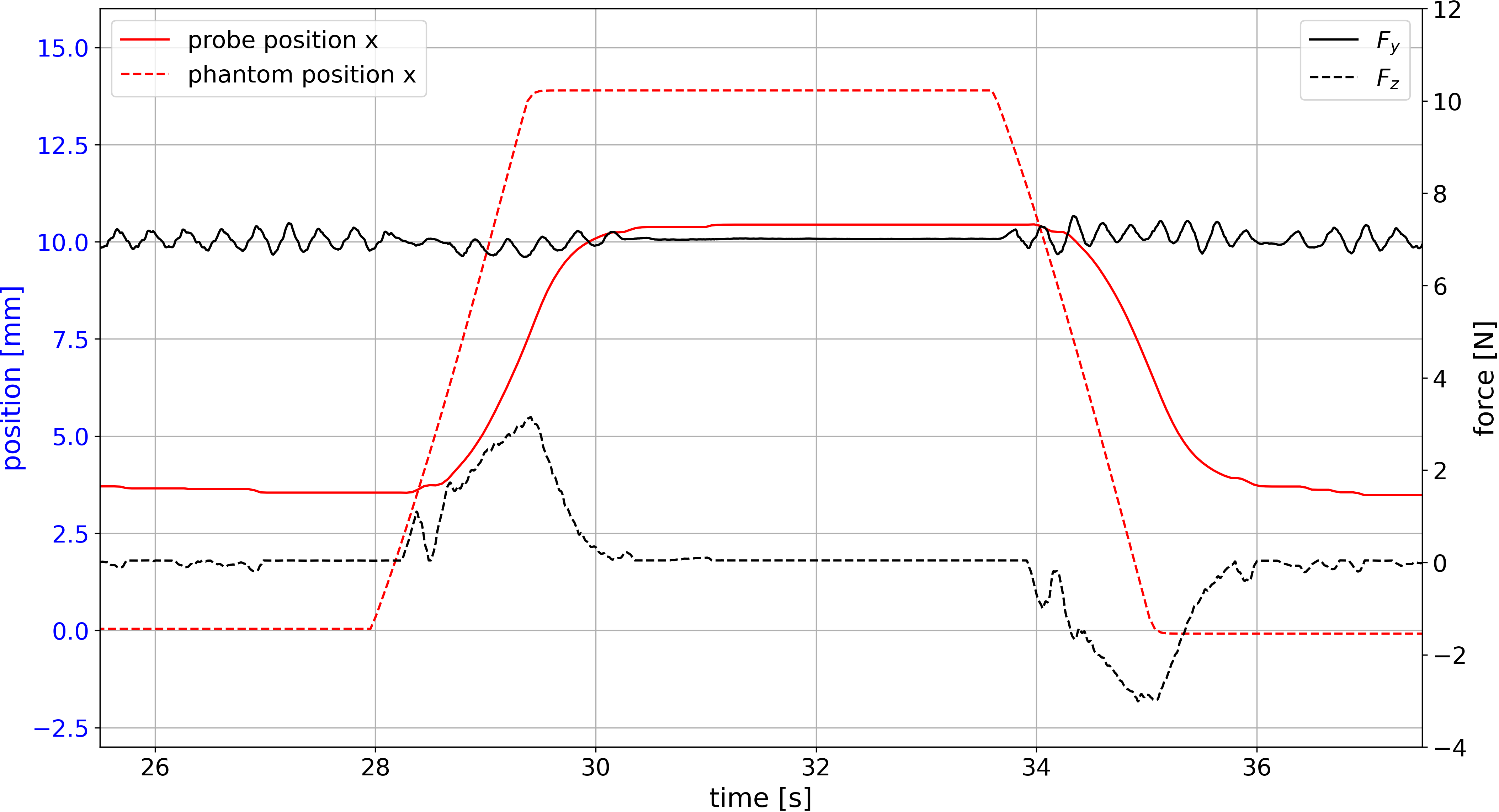}\label{fig:hor_mot}}\\
    \subfloat[\centering]{\includegraphics[width=\linewidth]{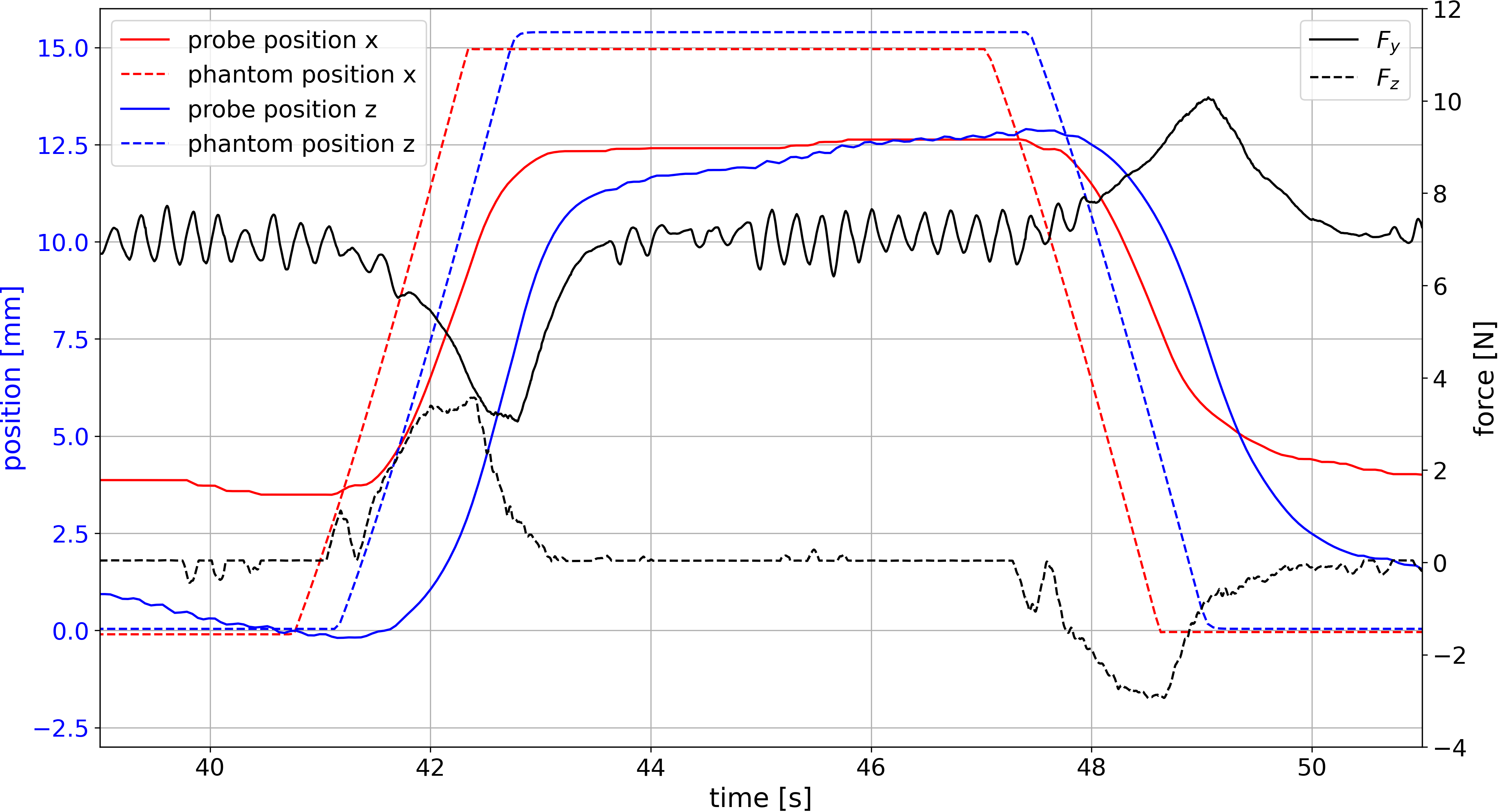}\label{fig:comb_mot}}
    \caption{Graphs of motion compensation: (a) vertical (V), (b) horizontal (H) and (c) combined (C) motion of the phantom }
    \label{fig:Graphs}
\end{figure}

The Figure \ref{fig:Graphs} shows the performance of motion tracking for three different motion scenarios: vertical movement, horizontal movement, and a combination of both movements. The position data in the graph represent the position of the US probe and the prostate phantom relative to their initial position, expressed in the world coordinate frame ($O_W$ in Figure \ref{fig:setup}). The force expressed in the graph is an external force acting on the US probe expressed in the probe's coordinate frame ($O_{TCP}$ in Figure \ref{fig:setup}).

The robot follows the movement with a time delay close to 1/2 second. This is due to the soft material of the phantom, which takes time to deform and exert a considerable force on the US probe.

For both vertical and more significantly horizontal movements, the ultrasound (US) probe performs smaller movement than the phantom, due to the controller's implemented deadzone. The greater offset in the horizontal direction can be explained by a loose fit between the probe and the phantom. During vertical movements, the probe is actively pressed against the phantom's surface, ensuring contact. With horizontal movements, on the other hand, the lack of active force leads to a delay before the phantom applies force on the probe after a change of direction.


\subsection{Prostate reconstruction}

System validation is performed by registering prostate point cloud representations recorded during various movements with those obtained in a stationary state. On average, each point cloud contains approximately 500,000 points, with no substantial differences in number of points observed between point clouds acquired under stationary conditions and those recorded during prostate motion. For registration, we use the Iterative Closest Point (ICP) algorithm, a well-established algorithm for aligning point clouds. To evaluate the ICP results, we use various metrics that quantify the precision of the transformation between two point clouds. One of these metrics is the \textit{Hausdorff distance}, defined as the maximum distance of a point in one point cloud to the closest point in the other. We also analyze the registration \textit{fitness value}, defined as the ratio of inlier correspondences to the total number of points, as well as the RMSE, which represents the average registration error over inlier correspondences.

\begin{figure}[ht]
    \centering
    \subfloat[\centering]{\includegraphics[width=0.49\linewidth]{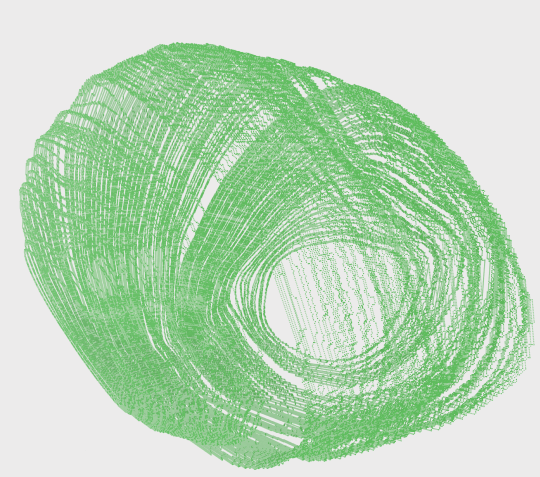}}
    \hspace{1px}
    \subfloat[\centering]{\includegraphics[width=0.49\linewidth]{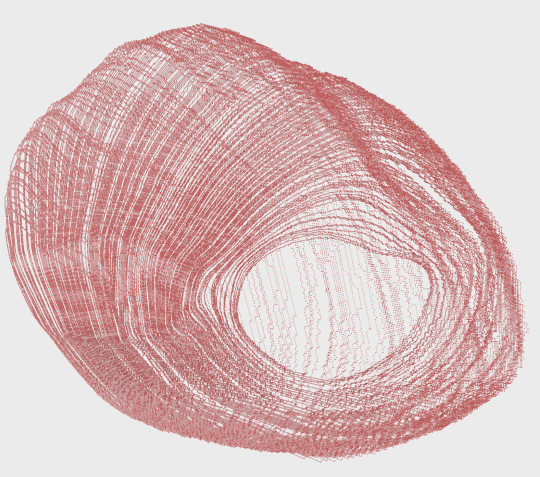}}\\
    \subfloat[\centering \label{fig:icp_c}]{\includegraphics[width=\linewidth]{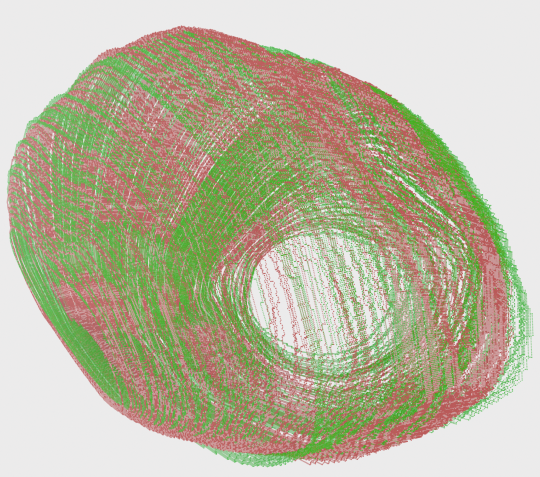}}
    \caption{(a) Source PC (Vertical Movement) (b) target PC (Stationary) (c) Result of ICP transformation. The green point cloud is the transformed source, and the red is the target point cloud. Example shown in the figure resulted in 5.916mm Hausdorff distance, 91.22\% fitness value and 0.41 mm RMSE, with a threshold value of 1 mm.  }
    \label{fig:icp}
\end{figure}

Figure \ref{fig:icp} illustrates one example of the ICP registration. The green points in Figure \ref{fig:icp_c} represent the transformed source point cloud, recorded while the phantom was moving vertically, and the red points represent the target point cloud, recorded while the phantom remained stationary. Although point clouds are different due to prostate movement simulated during one of the sweeps, ICP registration using the 1 mm threshold, which is defined as the largest distance between two points for them to be considered a pair during registration, results in 91.22\% fitness value and 0.41 RMSE value. In tables \ref{Tab:haus} and \ref{Tab:icp_table}, we give the mean values and standard deviations from our experiments. Table \ref{Tab:haus} shows that intra-set registrations (S–S) achieve the lowest mean value, indicating that point clouds acquired under stationary conditions are highly consistent. Since the Hausdorff distance reflects the largest discrepancy between paired points, this relatively low value demonstrates good alignment quality. In contrast, inter-set registrations (S–H, S–V, S–C) yield larger distances, as expected due to motion-induced variability.

\begin{table}[ht]
\centering
\begin{tabular}{@{}lcc@{}}
\toprule
\textbf{set pair} & \textbf{Hausdorff distance [mm]} \\ 
    \midrule
    S-S & $2.682 \pm 1.39$ \\ 
    S-H & $4.649 \pm 1.36$ \\ 
    S-V & $4.451 \pm 1.12$ \\ 
    S-C & $5.079 \pm 1.56$ \\ 
    \bottomrule
\end{tabular}
\caption{Mean $\pm$ standard deviation of Hausdorff distance over all pairwise registrations between point clouds from different sets (S: stationary set, H: horizontal, V: vertical, C: combined). Each set has 30 measurements.}
\label{Tab:haus}
\end{table}

The results from Table \ref{Tab:icp_table} show that intra-set registrations (S-S) performed over unique pairs of different point clouds consistently yield higher fitness and lower RMSE values compared to inter-set registrations (S-H, S-V, S-C). This outcome is expected and confirms that the scanning workflow is stable and produces reproducible data with only minor deviations. Across all thresholds, inter-set registrations follow a similar trend, indicating that the reduced ICP performance is systematic and not specific to a particular type of motion. Increasing the correspondence threshold improves the fitness value, but at the cost of higher RMSE, highlighting a trade-off between registration robustness and accuracy. In this context, thresholds around 0.6-0.8 mm appear to offer a good balance. These findings suggest that ICP can robustly handle small intra-set variations but becomes more challenged when aligning point clouds across different movement scenarios, where both metrics degrade.

\begin{table}[ht]
\centering
\begin{tabular}{@{}lccc@{}}
\toprule
\textbf{set pair} & \textbf{fitness}  & \textbf{RMSE [mm]} \\ 
    \midrule
    \textbf{Threshold 0.4 mm}\\
    S-S & $0.882 \pm 0.16$ & $0.175 \pm 0.02$ \\ 
    S-H & $0.431 \pm 0.16$ & $0.220 \pm 0.01$ \\ 
    S-V & $0.462 \pm 0.14$ & $0.226 \pm 0.01$ \\ 
    S-C & $0.473 \pm 0.14$ & $0.225 \pm 0.01$ \\ 
    \midrule
    \textbf{Threshold 0.6 mm}\\
    S-S & $0.965 \pm 0.08$ & $0.196 \pm 0.04$ \\ 
    S-H & $0.672 \pm 0.18$ & $0.297 \pm 0.02$ \\ 
    S-V & $0.693 \pm 0.13$ & $0.307 \pm 0.02$ \\ 
    S-C & $0.673 \pm 0.11$ & $0.305 \pm 0.02$ \\ 
    \midrule
    \textbf{Threshold 0.8 mm}\\
    S-S & $0.988 \pm 0.03$ & $0.203 \pm 0.05$ \\ 
    S-H & $0.832 \pm 0.16$ & $0.349 \pm 0.03$ \\ 
    S-V & $0.841 \pm 0.09$ & $0.368 \pm 0.03$ \\ 
    S-C & $0.794 \pm 0.09$ & $0.372 \pm 0.02$ \\ 
    \midrule
    \textbf{Threshold 1 mm}\\
    S-S & $0.994 \pm 0.01$ & $0.207 \pm 0.05$ \\ 
    S-H & $0.903 \pm 0.13$ & $0.380 \pm 0.05$ \\ 
    S-V & $0.921 \pm 0.05$ & $0.415 \pm 0.04$ \\ 
    S-C & $0.869 \pm 0.07$ & $0.430 \pm 0.03$ \\ 
    \midrule
    \textbf{Threshold 1.2 mm}\\
    S-S & $0.996 \pm 0.01$ & $0.209 \pm 0.05$ \\ 
    S-H & $0.936 \pm 0.11$ & $0.404 \pm 0.06$ \\ 
    S-V & $0.963 \pm 0.03$ & $0.453 \pm 0.05$ \\ 
    S-C & $0.915 \pm 0.05$ & $0.478 \pm 0.03$ \\ 
    \bottomrule
\end{tabular}
\caption{Mean $\pm$ standard deviation of ICP registration results over all pairwise registrations between point clouds from different sets (S: stationary set, H: horizontal, V: vertical, C: combined), for different correspondence thresholds. Each set has 30 measurements.}
\label{Tab:icp_table}
\end{table}

%% file: Text/4_Conclusion.tex
\section{Conclusion}
This paper presents a prototype robotic setup for motion compensation during prostate biopsies. Our system successfully tracks movements and scans the prostate to create a 3D reconstructed point cloud. Our work has certain limitations. First, we only use one medical phantom, which is a simplified representation of the prostate, making the segmentation process much easier. We intend to expand our work with more complex prostate phantoms and further develop our system for future clinical testing, which would allow for more realistic results.

To address the inadequacies in movement tracking performance, we plan to include a force-torque sensor on the robot's flange to eliminate the errors coming from the imperfect estimation from joint torques. This addition would make the setup bulkier, but the impact of the improved measurements on the robot's performance should justify it. The performance could be further improved by implementing a non-linear control law that allows for a faster response at high contact force while preventing the oscillations of the noisy force-torque signal.

Furthermore, future work will include the use of both elements of the ultrasound probe, linear and convex. This addition could provide more information about the movement and thus allow for better compensation, especially of movement and rotation around the x-axis of the probe. We have already started to develop this function, but have not been able to find a feature that is consistent between several phantoms and the real prostate.
